\PassOptionsToPackage{table,xcdraw}{xcolor}
\documentclass[sigconf]{acmart}
\usepackage{multirow}
\usepackage{pifont}
\usepackage{float} 
\usepackage{graphicx}
\usepackage{subfigure}
\usepackage{algorithm}
\usepackage{algorithmic}
\usepackage{svg}
\usepackage{listings}
\usepackage[most]{tcolorbox}

\usepackage{fancyhdr,arydshln}
\definecolor{mypink1}{rgb}{0.858, 0.188, 0.478}
\definecolor{green}{rgb}{0.158, 0.8, 0.3}
\definecolor{blue}{rgb}{0, 0.3, 0.8}

\definecolor{newcolor}{rgb}{.8,.349,.1}

\renewcommand\footnotetextcopyrightpermission[1]{}

\begin{document}

\title{VideoMind: An Omni-Modal Video Dataset with Intent Grounding for Deep-Cognitive Video Understanding}

\author{Baoyao Yang}
 \affiliation{ Guangdong University of Technology\country{China}
   % \institution{Paper under double-blind review}
 }

 \author{Wanyun Li}
 \affiliation{Wechat, Tencent\country{China}}

 \author{Dixin Chen}
 \affiliation{Guangdong University of Technology\country{China}}
 
\author{Junxiang Chen\normalsize{$^*$}}
 \affiliation{Wechat, Tencent\country{China} }

% \author{Anonymous authors}
%  \affiliation{
%    \institution{Paper under double-blind review}
%  }

  \author{Wenbin Yao}
 \affiliation{Wechat, Tencent\country{China}}
\authornote{
Corresponding author: JX Chen and WB Yao \textit{(\{caryjxchen,wenbinyao\}@tencent.com)} \\ \textit{This work has been submitted to a conference/journal for possible publication. Copyright may be transferred without notice, after which this version may no longer be accessible.}}

   \author{Haifeng Lin}
 \affiliation{Guangdong University of Technology\country{China}}

%%
%% By default, the full list of authors will be used in the page
%% headers. Often, this list is too long, and will overlap
%% other information printed in the page headers. This command allows
%% the author to define a more concise list
%% of authors' names for this purpose.
\renewcommand{\shortauthors}{BY Yang et al.}

%%
%% The abstract is a short summary of the work to be presented in the
%% article.
\begin{abstract}
This paper introduces \textbf{VideoMind}, a video-centric omni-modal dataset, which enables the deep cognition of video content and enhances feature representations of multi-modal data. The \textbf{VideoMind} dataset contains 103K video samples \textit{(3K for test only)}, each of which is accompanied by audio, as well as systematic and detailed textual descriptions. Specifically, every video sample, together with its audio data, is described across three hierarchical layers \textit{(factual, abstract, and intent)}, progressing from the superficial to the profound. 
In total, more than 22 million words are included, with an average of approximately 225 words per sample.
Compared with existing video-centric datasets, the distinguishing feature of \textbf{VideoMind} lies in providing intent expressions that are intuitively unattainable and must be speculated through the integration of context across the entire video.
The Chain-of-Thought (COT) text generation manner is introduced, wherein the mLLM is prompted to derive deep-cognitive expressions under step-by-step guidance. 
Upon the detailed descriptions, various annotations, including subject, place, time, event, action, and intent, are marked, serving a series of downstream recognition tasks.
%Additionally, we introduce a new benchmark, \textbf{DeME}, which is trained by 100K samples of the \textbf{VideoMind} to extract high-quality omni-modal embeddings. The remaining data (10K) were meticulously validated by expert annotators, establishing a gold-standard dataset for the evaluation of deep-cognitive video understanding.
%\textbf{DeME} has demonstrated outstanding performance on board cognitive-related applications, such as emotion recognition, intent recognition, \textit{etc}.
More crucially, we establish a gold-standard benchmark comprising 3,000 meticulously manual-validated samples for the evaluation of deep-cognitive video understanding. To more appropriately evaluate the model's deep understanding of videos, hybrid-cognitive retrieval experiments are designed, scoring by multi-level retrieval metrics. Evaluation results of multiple standard foundation models \textit{(InternVideo, VAST, UMT-L, \textit{etc.})} have been released.
It is believed that \textbf{VideoMind} is a powerful benchmark that facilitates fine-grained cross-modal alignment, but also promotes fields that need in-depth understanding of videos, such as emotion and intent recognition.
% 3000 samples of the \textbf{VideoMind} were meticulously validated by expert annotators, establishing a gold-standard dataset for the evaluation of deep-cognitive video understanding.
Data are publicly available on three hosting platforms \textit{(GitHub, Huggingface and Opendatalab)},
\textcolor{blue}{\textit{{https://github.com/cdx-cindy/VideoMind}}}.
\end{abstract}

\begin{teaserfigure}
\centering
\includegraphics[width=0.93\linewidth]{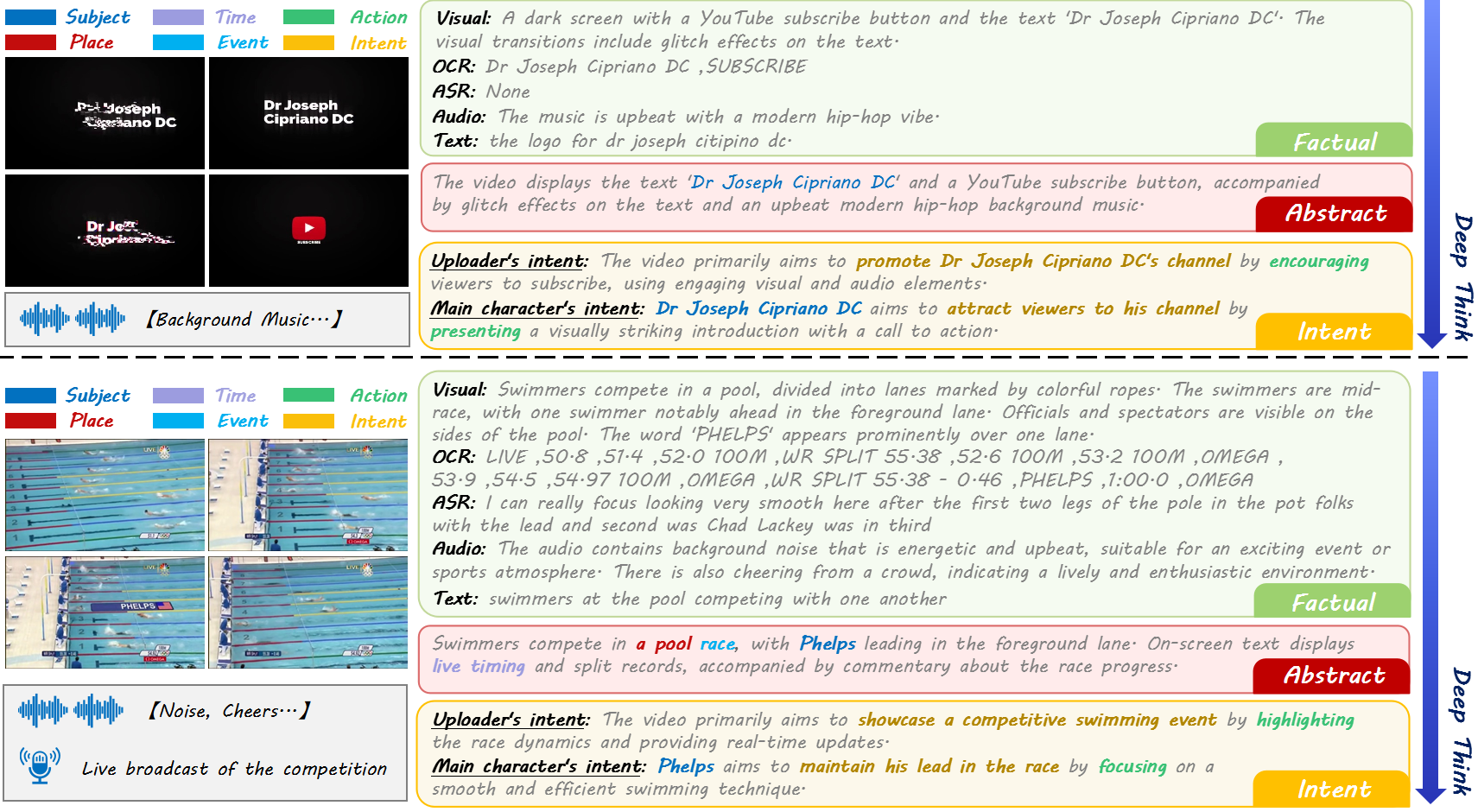}
 \vspace{-1em}
\caption{Examples of video clips in \textbf{VideoMind} \textit{(More examples are presented on \textcolor{blue}{\textit{{https://opendatalab.com/Dixin/VideoMind}}})}.}
\label{fig:teaser}
\vspace{1em}
\end{teaserfigure}

\maketitle

% \begin{figure*}[th]
%     \centering
%     \includegraphics[width=0.9\linewidth]{figures/samples.png}
%     \vspace{-0.5em}
%     \caption{Examples of video clips in VideoMind \textit{(More examples are presented on \textcolor{blue}{\textit{{https://opendatalab.com/Dixin/VideoMind}}})}.}
%     \label{fig:example}
% \end{figure*}

\begin{table*}[th]
\caption{Comparisons between the proposed \textbf{VideoMind} and current video-centric multi-modal datasets}
\label{tab:stat}
\vspace{-1em}
    \begin{tabular}{lllcccccccccc}
    \hline
      Dataset   &  Domain  &  \#Clips &Text Source & Len$_{text}$  & Image &  Video & Audio & ASR & OCR & Tag & Intent\\
    \hline
    MSR-VTT \cite{xu2016msr} & Open & 10K  & Manual & 9.3 & &\ding{51}\\
    MSVD \cite{chen2011collecting} & Open & 1970  & Manual & 8.7 & &\ding{51}\\
    LSMDC \cite{rohrbach2017movie} & movie & 118K  & Manual & 7.0 & &\ding{51} \\
    ANet Caption \cite{krishna2017dense} & Action & 20K & Manual & 13.5 & &\ding{51}\\
    VaTex \cite{Wang_2019_ICCV} & Open & 41.3K  & Manual & 15.2 & &\ding{51}\\
    HT100M \cite{miech2019howto100m} & Instructional & 136M  & ASR & 4 & &\ding{51} & &\ding{51} \\
    HD-VILA-100M \cite{xue2022advancing} & Open & 103M  & ASR & 32.5 & &\ding{51} & &\ding{51} \\
    WebVid-10M \cite{bain2021frozen} & Open & 10.7M  & Alt-texts & 12 & &\ding{51} \\
    How2 \cite{sanabria18how2} & instructional & 80K & Manual & 20 & &\ding{51} \\
    VALOR \cite{liu2024valor} & Open & 1.1M & Manual & 16.4 & &\ding{51} &\ding{51} \\
    VAST \cite{chen2024vast} & Open & 27M & Generated & 32.4 & &\ding{51}&\ding{51} \\
   InternVid \cite{wang2023internvid} & Open & 234M  & Generated & 17.6 & &\ding{51} & &\ding{51} & &\ding{51} \\
   GME \cite{zhang2024gme} & Open & 1.1M  & Generated & -- &\ding{51}\\
        \textbf{VideoMind} \textit{(Ours)} & Open & 103K & COT-Generated & 225 &\ding{51}&\ding{51}&\ding{51}& \ding{51} & \ding{51} & \ding{51} & \ding{51}\\
    \hline
    \end{tabular}
\end{table*}

\section{Introduction}
\label{sec:introduction}

Video has become the predominant medium for message dissemination with the advancement of social media. Precise understanding of video content, particularly its underlying purpose and intent, is essential for promoting intelligent communication and safeguarding the legality and security of the internet.
Recently, dozens of multi-modal foundation models, such as Video-LLaVA \cite{lin2024video}, SoRA, and Qwen-vl~\cite{wang2024qwen2}, have been published. 
They have been reported to achieve impressive performance in generation and conversation tasks. 
Their success is heavily contingent upon large-scale pre-training with multimodal data. Although many large enterprises have privatized pre-training data, quite a few video-centric databases have still been released. 
For instance, the Max Planck Institute for Informatics, in collaboration with UC Berkeley, published the LSMDC \cite{rohrbach2017movie} dataset, which comprises over 100K movie clips annotated with manual descriptions of the video content.
The Max Brain then proposes a 10M dataset, WebVid \cite{bain2021frozen}, wherein the textual descriptions of videos are sourced from the web.
HD-VILA-100M~\cite{xue2022advancing} further expands the data scale to 100M, and the video clips are automatically annotated with their ASR results. 
Dissatisfied with hasty text extraction, Shanghai AI Lab adopt mLLM-based text generation techniques to enhance the quality of textual expression \cite{wang2023internvid}.
Subsequent studies, such as VAST \cite{chen2024vast} and VALOR \cite{liu2024valor}, have further incorporated audio information to enrich the range of data modalities and enhance data diversity. This integration broadens the receptive field of multi-modal large language models (mLLM) and extends their application scope.
However, despite their efforts in providing rich information for model training, the existing datasets exhibit the following limitations:

\textit{\textbf{1) Overly concise text expression}}: Although multi-modal elements are involved in the existing datasets, the video content is typically under-described using a single brief sentence of around 20 words. This textual description often corresponds to only a subset of video frames or specific regions, resulting in a significant discrepancy of information across modalities.

\textit{\textbf{2) Lack of in-depth interpretation}}: The descriptions of videos remain at the level of purely visual observation, offering no additional underlying information that requires in-depth thinking and reasoning. This causes the model to have no idea about the underlying intent of videos.

\textit{\textbf{3) Heavily task bias}}: Most existing datasets are specifically curated for particular tasks, like captioning or video Q\&A. Such data may guide the model to extract embeddings that facilitate text generation, yet neglect general representativeness.

The above issues limit the generalization of foundation models, more importantly, hinder a deeper cognitive understanding of video content.
The lack of in-depth understanding of video content will impede the accurate alignment of user demands, ultimately leading to a loss of commercial value. Furthermore, in-depth video comprehension is essential for maintaining order on social platforms, enabling the effective identification and suppression of low-quality content while safeguarding copyright integrity.
To this end, this work proposes a new dataset, \textbf{VideoMind}, which provides comprehensive \textit{(broad and depth)} textual interpretation of video content.
As illustrated in Figure \ref{fig:teaser}, each video sample is systematically described at three hierarchical layers—factual, abstract, and intent—ranging from surface-level to in-depth interpretations. Specifically, these descriptions are generated through a step-by-step Chain-of-Thought (COT) prompting approach applied to an mLLM, the overall pipeline of which is detailed in Section \ref{sec:text-gen}.
At the core of the intent layer, we establish a clear rule for intent expression and design two role-playing tasks aimed at minimizing ambiguous expressions and ensuring accurate speculation of the video's underlying intent.
Additionally, \textbf{VideoMind} encompasses omni-modal data (images, videos, audio, and text) along with a wide range of associated information, such as automatic speech recognition (ASR), optical character recognition (OCR), and various semantic tags, as summarized in Table \ref{tab:stat}.
%This paper also develops an omin-modal model named \textbf{DeME}, which is pre-trained by the \textbf{VideoMind} data.
%\textbf{DeME} is a generic multi-modal embedder, and the effectiveness of its embeddings has been validated through various intent-related downstream tasks, including deep-cognitive cross-modal retrieval, emotion recognition, and intent recognition, \textit{etc.}

In summary, this paper introduces the FIRST deep-cognitive omni-modal video dataset, \textbf{VideoMind}, which encompasses comprehensive descriptions of video content and detailed explanations of underlying intent.
By enhancing the intent-aware capabilities of data embeddings, \textbf{VideoMind} enables omni-modal foundation models to thoroughly explore intrinsic relations across samples, thereby improving precise matching and facilitating effective quality monitoring in social network.
%\item  We train an omin-modal model, named \textbf{DeME}, based on the \textbf{VideoMind} data. \textbf{DeME} is a generic embedder framework, which owns has high reusability and extensibility. 
%\textbf{DeME} exhibites superior performance across a range of general video recognition tasks and intent-perceptive scenarios, notably surpassing the SOTA foundation model by xx\% in zero-shot video-text retrieval.

%\end{itemize}

\section{Related Work}

%\subsection{Video-centric Multi-modal datasets}
Video-text pairs are the foundation of training powerful multimodal models, enhancing video understanding and deriving cross-modal representations to support various downstream tasks.
Multiple centers have put in enormous effort on data acquisition, establishing video-centric datasets \cite{xu2016msr,chen2011collecting,rohrbach2017movie,krishna2017dense,Wang_2019_ICCV,sanabria18how2,liu2024valor} that rely on manual labeling.
As labor-intensive projects, these datasets typically exhibit limitations in data volume or suffer from overly simplified textual representations. For instance, MSVD \cite{xu2016msr} comprises only 1,970 video-text pairs, while the samples in LSMDC \cite{krishna2017dense} are, on average, expressed using seven words.
To reduce the labor cost of writing, other works leverage Alt-text \cite{bain2021frozen,thomee2016yfcc100m,hu2022scaling} in web images and ASR \cite{miech2019howto100m,zellers2021merlot,xue2022advancing} in video clips as textual descriptions of the video content. This strategy provides great convenience for data collection, consequently, the dataset scale has been enlarged to over 100 million. However, the representativeness of Alt-text and ASR outputs is debatable, as some videos may lack audio speech altogether, and the quality of Alt-text cannot be consistently ensured.

% Large-scale visual-text data pairs

% can effectively promote cross-modal learning, and high-quality video-centric multimodal datasets are necessary for video understanding tasks and learning powerful video-text representations.

% Initially, researches produced video-centric data relied on manual labeling \cite{xu2016msr,chen2011collecting,rohrbach2017movie,krishna2017dense,Wang_2019_ICCV,sanabria18how2,liu2024valor}.

% To learn robust model, existing works usually leverage Alt-text \cite{bain2021frozen,thomee2016yfcc100m,hu2022scaling} in web images and ASR \cite{miech2019howto100m,xue2022advancing} in video clips for large‌-scale and effective‌ dataset. 

Nowadays, the advancement of mLLM has catalyzed a paradigm shift toward generative approaches \cite{chen2024vast,wang2023internvid,zhang2024gme} for dataset construction. 
In recognition of the powerful data parsing capabilities of mLLM, Shanghai AI Lab releases a large-scale video-text dataset, named InternVid  \cite{wang2023internvid}, in which the mLLM is requested to generate a simple textual description for every video clip.
Later on, the mLLM-derivation strategy is extended to tri-modality, a large-scale vision-audio-language dataset named VAST \cite{chen2024vast} is published. This starts the process of omni-modal representation learning.  However, the disproportionate prioritization of dataset scale in these datasets has resulted in the inclusion of low-quality videos and ambiguous captions.
Rather than merely pursuing scalability, Nanjing University emphasizes enhancing the quality and aesthetic value of videos. To this end, they propose OpenViD-1M \cite{nan2024openvid}, a dataset comprising 0.4 million videos in 1080P resolution.

Although dozens of video-centric datasets have been proposed, few studies consider the comprehensiveness of text expressions. A substantial gap in information content remains indelible across different modalities, leading to natural obstacles for omni-modal learning.
Recent works, such as Video-MME \cite{fu2025video} and IntentQA \cite{li2023intentqa}, have attempted to supplement the text content through a Q\&A manner. However, the descriptions tend to be one-sided, focusing specifically on certain positions or frames within a video.
Therefore, this work aims to address this gap, providing more comprehensive and in-depth textual annotations for videos, thereby facilitating video understanding and enhancing omni-modal representations.

\begin{figure*}[t]
    \centering
    \includegraphics[width=0.75\linewidth]{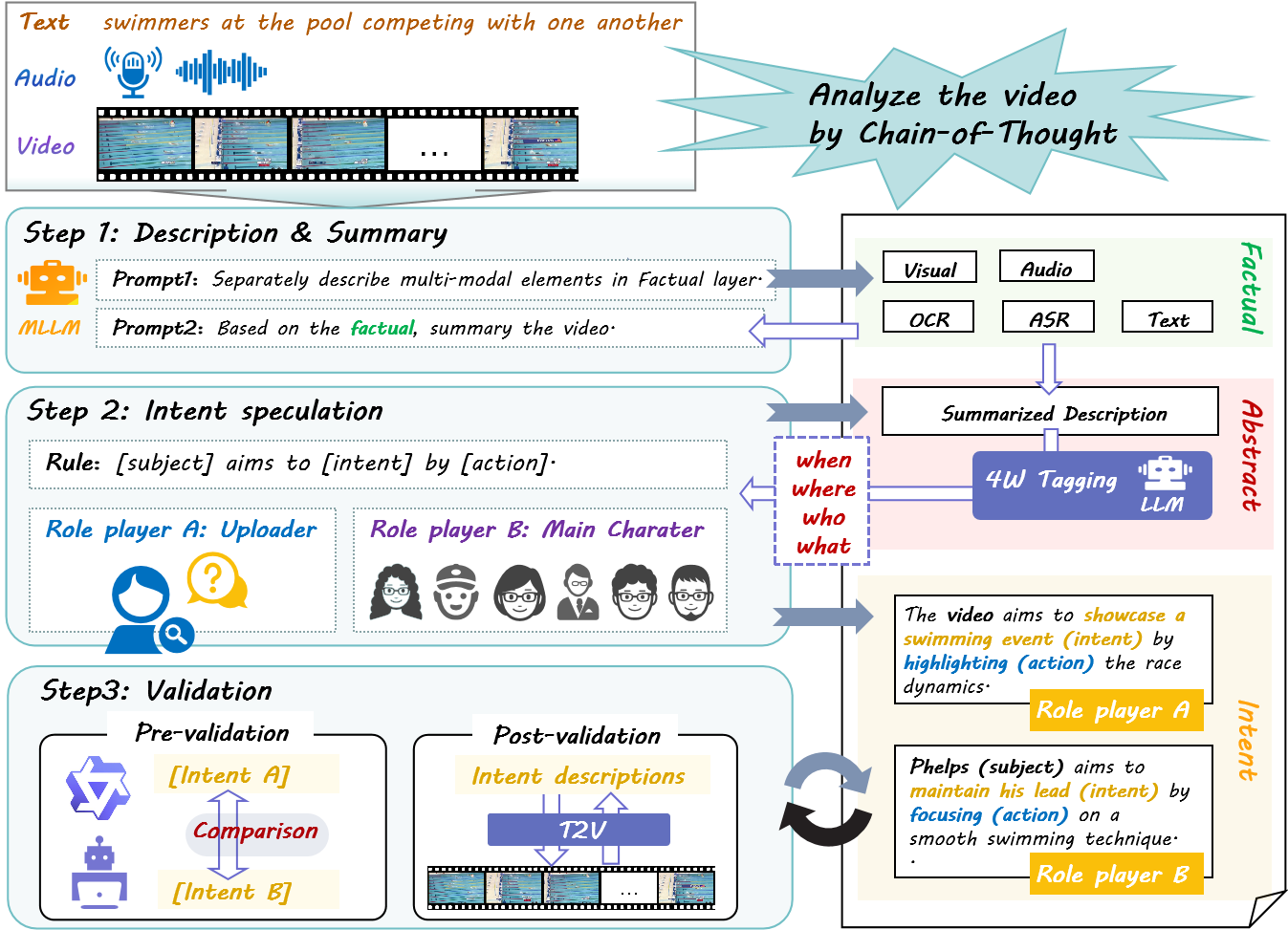}
    \vspace{-0.5em}
    \caption{Text generation pipeline of \textbf{VideoMind}.}
    \label{fig:pipeline}
\end{figure*}

\section{VideoMind: An Omni-modal Dataset with Broad and Deep Grounding}
\label{sec:videomind}
A comprehensive and in-depth omni-modal dataset serves as a critical foundation for bridging the gap in mLLM's capability to understand implicit information, such as intent, emotion, and motivation. 
For a comprehensive understanding of videos, \textbf{VideoMind} provides full-sided modalities \textit{(including key frames, video, audio, and text)}, as well as deep-thinking text expressions spanning from what-you-see factual to in-depth intent speculation. Specifically, the text descriptions of every video include three layers: factual, abstract, and intent layers. The fact layer gives a comprehensive and detailed description of content from various modalities, while the abstract layer offers a summary of video samples based on factual descriptions. Leveraging the information from the above two layers, the intent layer further speculates and records the purpose of the video. In addition, video information is disassembled into non-overlapping elements: visual (regardless of OCR), audio (background sound), OCR, ASR, and text. A variety of tags, such as subject, place, and time, are annotated, supporting a broad spectrum of downstream tasks, such as event recognition, identification, and scene recognition.

In sum, \textbf{VideoMind} comprises 103,000 video samples paired with approximately 22 million textual annotations. The statistical comparison regarding the data volume, text length, information tag, \textit{etc.}, is given in Table \ref{tab:stat}.
In order to provide a high-quality and standardized evaluation for tasks involving an in-depth understanding of videos, 3,000 samples are carefully selected from the \textbf{VideoMind} dataset, establishing a benchmark validation set. The rationality of all samples in the validation set has been independently validated by three professional annotators.

\subsection{Data Curation}
To ensure the diversity of video intent, we select videos posted on social networks as our original resources.
Following InternVid~\cite{wang2023internvid}, clips in the \textbf{VideoMind} dataset are sourced from publicly available YouTube videos and encompass a wide range of domains, including gaming, news, entertainment, sports, \textit{etc.} \textit{(See \textbf{Section \ref{sec:stat}} for category statistics)}. Samples involving sensitive information, such as political tendencies and pornography, are excluded from consideration.
To ensure that samples contain sufficient intent information, all video clips are at least 5 seconds in duration. For each video clip, we randomly select 12 frames as representatives of the image modality.
The original audio and text corresponding to the video clips are simultaneously extracted, forming a quadruple expression \textit{(video, image, audio, text)} for each sample.
The selected quadruples would go through the text generation process (Figure \ref{fig:pipeline}), and only those pass the double-validation are retained.
The selected quadruples will undergo the text generation process (Figure \ref{fig:pipeline}), comprehensively and deeply rewriting the text content, and only those that successfully pass the dual-validation procedure will be retained in the \textbf{VideoMind} dataset.

\subsection{COT-based Text Generation}
\label{sec:text-gen}
Providing extensive and in-depth textual descriptions of videos represents the primary characteristic of \textbf{VideoMind}. This capability facilitates a deeper understanding of video content and enhances various downstream tasks related to cognition and emotion. To achieve this objective, an mLLM (Qwen2.5-Omni) is introduced and prompted to progressively generate multi-layer expressions of video content through three-stage parsing in \textbf{a COT manner}, as illustrated in Figure \ref{fig:pipeline}. The expressions in a deeper layer are always induced based on the analysis and speculation of the previous layers.
%\textit{(The prompt used for text generation is given in the Appendix)}.
Consequently, texts of videos are enriched, evolving from superficial presentation to deeper connotations: 1) Factual layer, which describes observable and audible elements; 2) Abstract layer, which synthesizes multi-modal information into a cohesive summary; 3) Intent layer, which speculates on the motivations of video creators and the primary subjects within the videos.
We elaborate in detail on each step of the generation process in the following.

\noindent \textbf{\textit{Step 1: Multiperspective Descriptions and High-Level Summary.}}
Unlike previous video-centric datasets, \textbf{VideoMind} hopes to fully reflect information from various sources without any loss. Therefore, it is required that the mLLM separately describe multimodal data in the first stage. As a result, five non-overlapping elements are recorded in the factual layer. That is, 1) Visual: Description of what you see regardless of graphic text;
2) Audio: Description of what you hear, regardless of human speech;
3) OCR: optical character recognition result of the video;
4) ASR: automatic speech recognition result of the audio;
5) Text: the raw text of the video, which is written manually.
In the factual layer, multi-modal data are detailed described by category. This facilitates bridging the cross-modal information gap, as visual and audio data are often more informative than the original texts corresponding to partial frames of the video.
Then, the mLLM is requested to summarize content covering all information in the factual layer, forming a brief overview in the abstract layer. 
The core elements, such as subject, place, time, and event, are also annotated in this layer \textit{(see Section \ref{sec:tag})}.

\noindent \textbf{\textit{Step 2: Intent Speculation.}}
According to the descriptions in the factual and abstract layers, we further prompt the mLLM to infer the intent of each video.
To prevent ambiguous or chaotic expressions that may interfere with the analysis, we have established a clear rule for intent expression. Following the natural law that intent always accompanies action, the expression rule of intent speculation is formulated as: \textit{[subject] aims to [intent] by [action]}. This prescribed expression facilitates the subsequent extraction and localization of intents.
Here, we take one more step to promote intent interpretation: two role-playing tasks are designed and conducted separately for intent speculation. Specifically, the role player A is the uploader of the video - the mLLM needs to deduce the purpose of uploading videos. The role player B, on the other hand, is the main character in the video. The mLLM imagines itself as the protagonist and explains the motivation behind the depicted action.
Through this role-playing mode, the development direction of the thinking chain has been clarified, presenting the video intent from multiple perspectives while effectively reducing the probability of model fantasy.

\begin{figure*}[t]
    % \centering
    \includegraphics[height=0.23\textwidth]{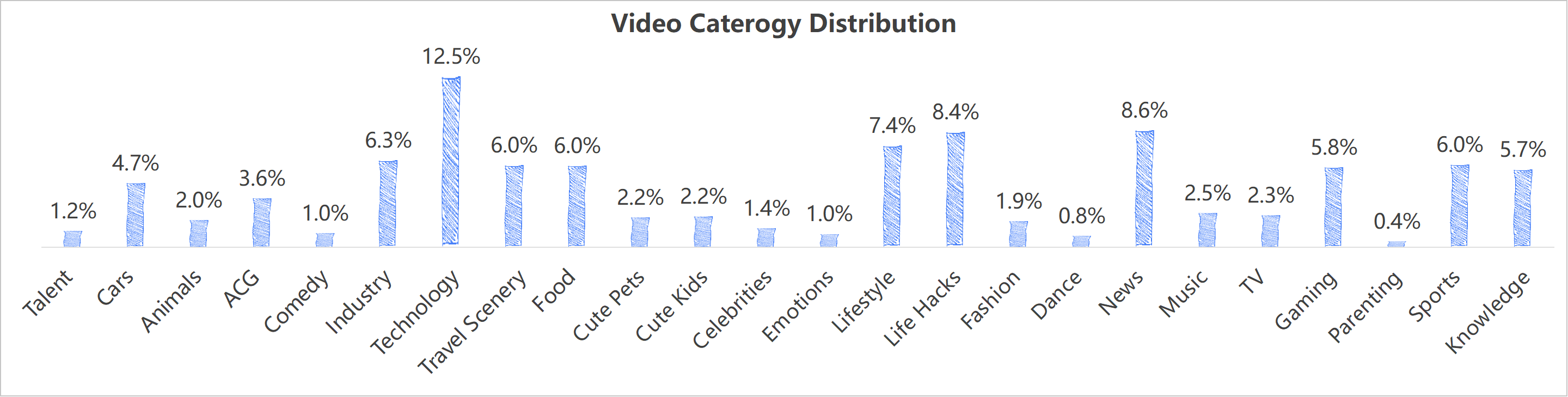} \\
    \vspace{0.5em}
    \includegraphics[height=0.17\textwidth]{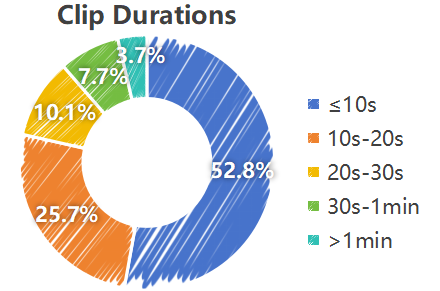}
    \hspace{2.5em}
    \includegraphics[height=0.17\textwidth]{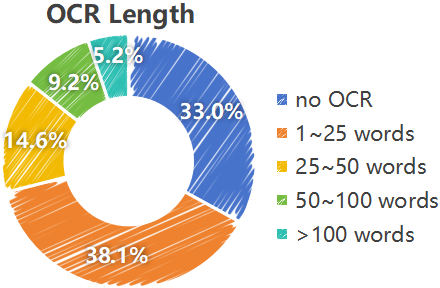} \hspace{2.5em}
    \includegraphics[height=0.17\textwidth]{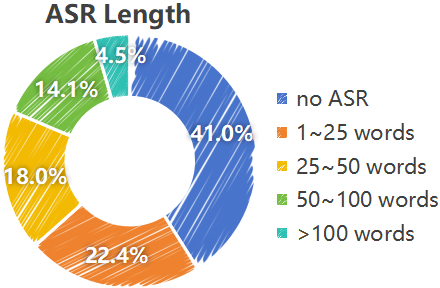}
    \vspace{-0.5em}
    \caption{Video statistics in \textbf{VideoMind}.}
    \label{fig:stat}
\end{figure*}

\noindent \textbf{\textit{Step 3: Validation: Quality Control.}}
The validation process is conducted to assess the reliability of the intent speculation results (descriptions at the intent layer) for both the video uploader and the main characters.
Each intent expression must undergo both pre-validation and post-validation processes to qualify as a valid sample in the \textbf{VideoMind} dataset.
During the pre-verification phase, an additional mLLM is incorporated to execute the identical intent speculation task.  This would yield two fixed-format intent expressions.
Following the specified format, the terms representing the intent in both texts can be easily extracted, and the intent speculation results are assessed by evaluating the embedding similarity of these two intent-revealed terms.   The intent is deemed correct only when the semantic meanings conveyed by the intent-expression words from both models exhibit similarity.
For post-validation, text-to-video generation techniques are employed. A 10-second video is produced for each intent expression using Wan2.1, and two expert annotators evaluate the reasonableness of the video content to implicitly assess the quality of the text writing.

Furthermore, we carefully selected 3,000 samples encompassing a broad spectrum of categories. The quality of the intent layer is rigorously evaluated by professional annotators. Subsequently, we established a benchmark serving as the first standardized criterion for the deep understanding of videos.

\subsection{Tagging}
\label{sec:tag}
\textbf{VideoMind} further delivers 6W-element tags, including subject (who), place (where), time (when), intent (why), action (how), and event (what), to expand its scope of application, supporting various downstream tasks.
Specifically, an LLM (Qwen2.5-vl) is tasked with identifying and highlighting nouns associated with location, time, and event from the expressions within the abstract layer, respectively.
Moreover, with fixed expression rules of intent, \textit{i.e.}, \textit{[subject] aims to [intent] by [action]}, subject, action, and intent are automatically annotated in the intent layer.
An example of the tagging results is presented in Figure \ref{fig:teaser}, wherein the 6W elements are distinguished by labels in various colors.

\begin{figure*}
    \centering
    \subfigure[Intent-related actions of uploaders]{\label{fig:wc-action-uploder}\includegraphics[width=0.23\textwidth]{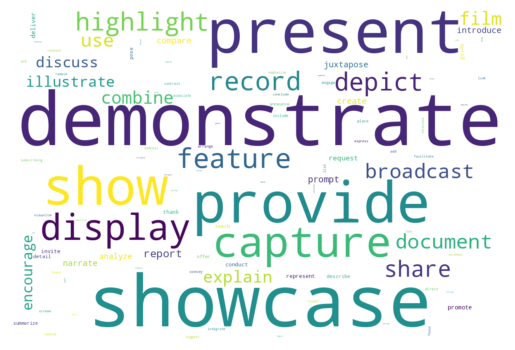}} \hspace{0.5em}
    \subfigure[Intent of uploaders]{\label{fig:wc-intent-uploder}\includegraphics[width=0.23\textwidth]{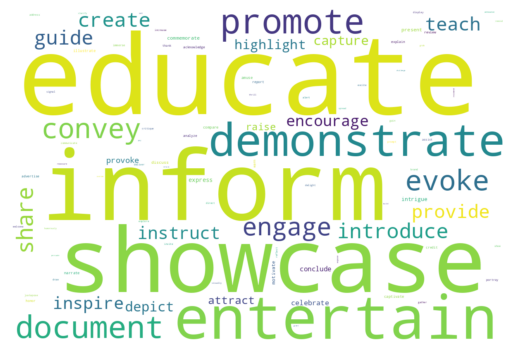}} \hspace{0.5em}
    \subfigure[Intent-related actions of characters]{\label{fig:wc-action-uploder}\includegraphics[width=0.23\textwidth]{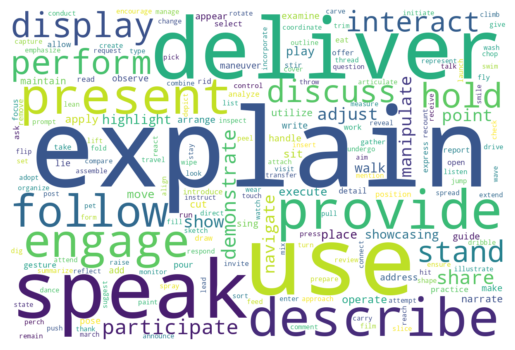}}    \hspace{0.5em}
    \subfigure[Intent of characters]{\label{fig:wc-intent-character}\includegraphics[width=0.22\textwidth]{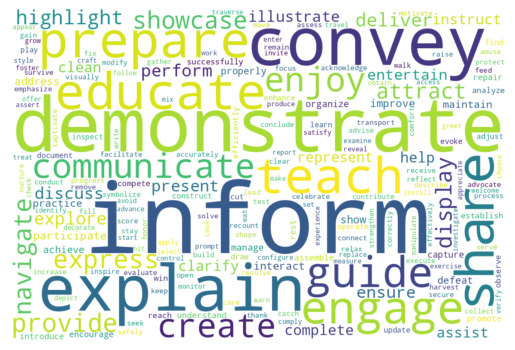}}
   \\
    \subfigure[Audio styple]{\label{fig:wc-audio}\includegraphics[width=0.23\textwidth]{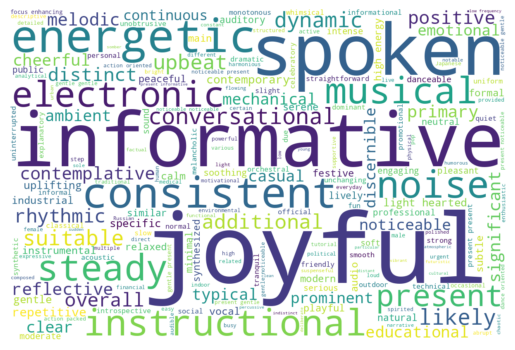}}  \hspace{1.5em}
    \subfigure[Subject]{\label{fig:wc-subject}\includegraphics[width=0.23\textwidth]{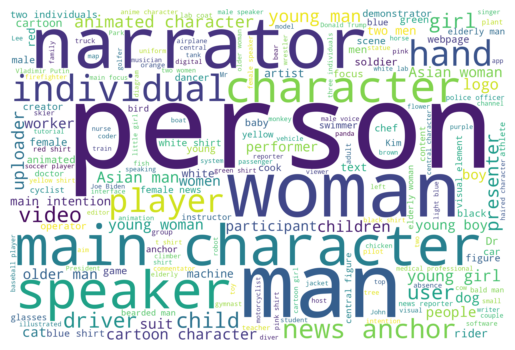}} \hspace{1.5em}
    \subfigure[Place]{\label{fig:wc-place}\includegraphics[width=0.23\textwidth]{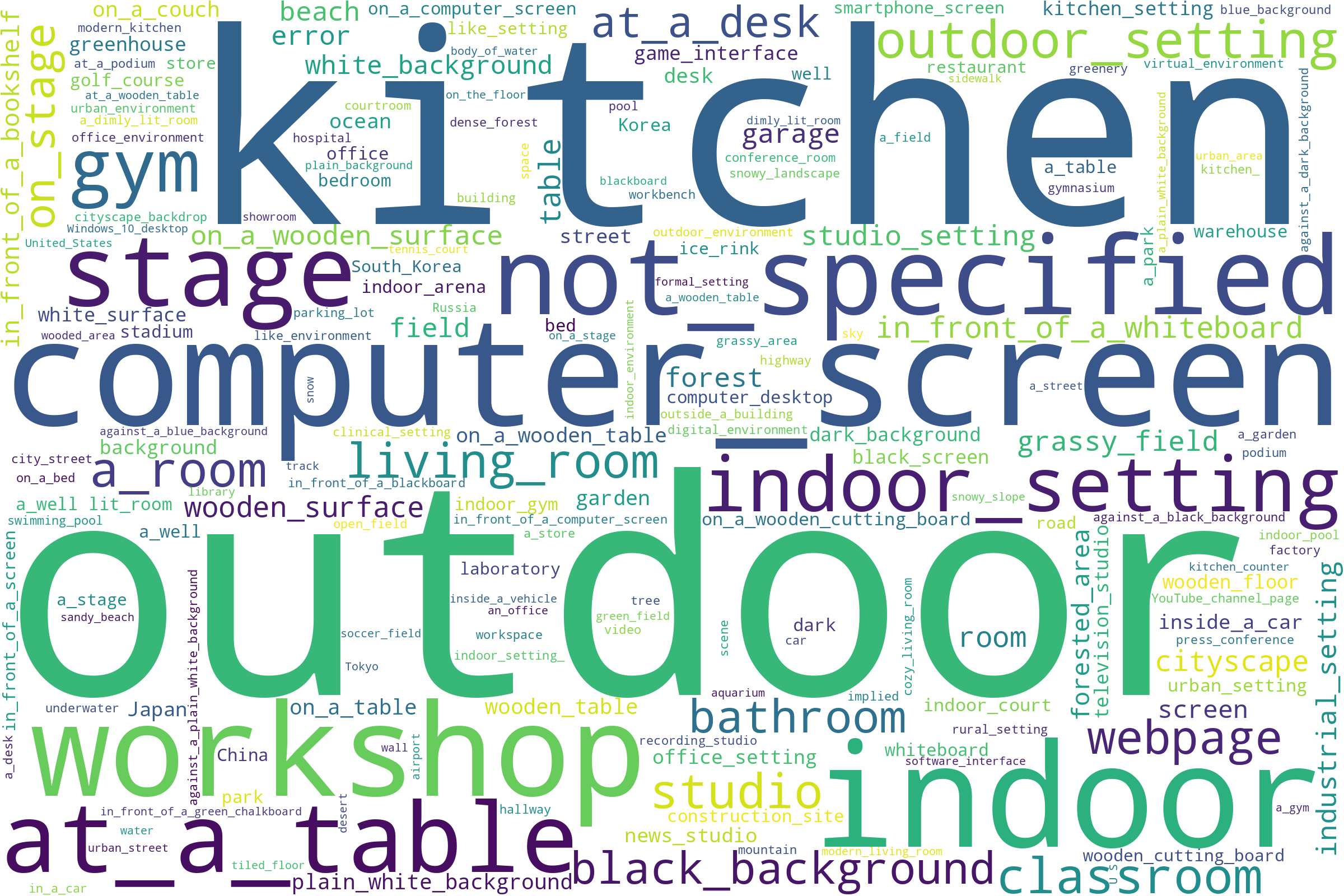}}  
    \vspace{-1em}
    \caption{The word cloud of intent, the intent-driven actions, audio style, subject, and place in the \textbf{VideoMind} dataset.}
    \label{fig:wordcloud}
\end{figure*}

\subsection{Statistics and Features}
\label{sec:stat}
The key statistics and features of \textbf{VideoMind} are summarized in Table \ref{tab:stat}, providing a comparison with existing video-centric multi-modal datasets.

\noindent \textbf{\textit{Data Diversity.}} 
To widely reflect the public's behavioral intentions, \textbf{VideoMind} includes various samples posted on social media, covering 45 countries/regions and 24 categories. 
The most common languages are English \textit{(EN, 39.7\%)},  Korean \textit{(KO, 9.0\%)}, Chinese \textit{(CN, 7.5\%)}, Japanese \textit{(JA, 7.5\%)} and Russian \textit{(RU, 3.8\%)}. A fair portion of the videos are language-free, corresponding to 14.7\% of samples with neither OCR nor ASR. 
The upper subfigures of Figure \ref{fig:stat} show that a considerable proportion of the videos do not have OCR (33.0\%) or ASR (41.0\%) output. For videos that include OCR/ASR, the majority of OCR/ASR outputs are shorter than 100 words in length. Only 5.2\% and 4.5\% of the videos contain long OCR and ASR outputs \textit{(>100 words)}, respectively.
The prevalent categories of videos include technology \textit{(12.5\%)},  news \textit{(8.6\%)}, life hacks \textit{(8.4\%)}, lifestyle \textit{(7.4\%)}, industry \textit{(6.3\%)}, as shown in Figure \ref{fig:stat}. In terms of duration, clips lasting fewer than 5 seconds are deemed invalid samples and excluded from \textbf{VideoMind}, as these excessively brief videos typically fail to encapsulate meaningful intent.
The average duration is 16.23 seconds per video.
52.8\% of the videos fall between 5 to 10 seconds, while 25.7\% lasts 10 to 20 seconds. Only a few samples (3.7\%) are over 1 minutes.

\noindent \textbf{\textit{Comprehensiveness and Richness (broad and deep).}} 
% Compared with the current datasets, the two primary characteristics of \textbf{VideoMind} lie in breadth and depth. \textbf{\textit{1) Breadth}}:
\textbf{VideoMind} is an omni-modal dataset, where each sample comprises presentations from various modalities, including key frames, audio, and text.
More importantly, the information on each modality is systematically categorized and subsequently described in detail. 
% Video information is disassembled into five non-overlapping elements: visual (regardless of OCR), audio (background sound), OCR, ASR, and text. 
The primary characteristic of \textbf{VideoMind} is its ability to provide extensive and in-depth descriptions of videos.
\textbf{VideoMind} generates comprehensive textual descriptions for videos in a COT manner, with an average of 225 descriptive words per sample, which is approximately \textit{\textbf{10 times}} that of existing video-centric datasets. Specifically, there are on average 143, 38, 43 words in the factual, abstract, and intent layers, respectively.
\textbf{VideoMind} also encompasses the most extensive range of modality information, including image, video, audio, ASR, OCR, and text, to the best of our knowledge. It also provides rich tags on character \textit{(who)}, place \textit{(where)}, time \textit{(when)}, event \textit{(what)}, action \textit{(how)}, and intention \textit{(why)}, supporting a large variety of downstream recognition tasks, such as cross-modal retrieval, intent VQA, intent recognition, and emotion recognition. Among these tags, character, place, time, and event are labeled in the abstract layer, whereas the action and intent are annotated in the intent layer. 

To demonstrate the intent expressions in \textbf{VideoMind}, we statistically analyze the word distribution of intents and intent-related actions performed by both video uploaders and the primary characters within the videos. From Figure \ref{fig:wordcloud}, it can be observed that video uploaders tend to \textit{demonstrate}, \textit{showcase}, \textit{present}, and \textit{provide} information to viewers, while the main characters commonly \textit{explain}, \textit{deliver}, and \textit{speak} their purpose.
Education, notification, showcase, and entertainment are the most common intent of uploaders; coincidentally, most of these are also the intent frequently presented by the main characters within the videos. This phenomenon of intent overlap is primarily attributed to the fact that the uploaders of a certain percentage of videos are identical to the main characters within these videos, particularly in categories such as lifestyle, technology, and talent.
In addition, the main characters demonstrate a greater range of potential intent compared to the uploaders, with which communicate, prepare, explain, and enjoy are additionally prominent in the word cloud representing the intent of the main characters (Figure \ref{fig:wc-intent-character}).

We also display the word clouds related to the audio style, subjects, and places that appear in the video in Figures \ref{fig:wc-audio}, \ref{fig:wc-subject}, and \ref{fig:wc-place}. 
According to the statistical results, the most commonly used style of background music in videos is joyful. Also, electronic, energetic, and musical are popular options as background music. Other videos, like sports or gaming, may be filled with the noisy voices of the crowded, showing audio styles of informative, spoken, and noise. 
Moreover, in these social media, there are more female subjects than male ones; the frequency of words like "woman" and "girl" is higher than that of male pronouns such as "man" and "boy". In terms of places, indoor/outdoor, kitchen, computer screen, workshop, studio and stage, are the most common video shooting venues.

\begin{table*}[t]
\caption{Results of hybrid-cognitive text-to-video retrieval on \textbf{VideoMind-3K}.}
\label{tab:t2v-deep-retrieval}
\vspace{-1em}
\small
\begin{tabular}{c|p{1.8em}p{1.8em}p{2.2em}p{3em}|p{1.8em}p{1.8em}p{2.2em}p{3em}|p{1.8em}p{1.8em}p{2.2em}p{3em}|p{1.8em}p{1.8em}p{2.2em}p{3em}}
\hline
   & \multicolumn{4}{c|}{factual$\rightarrow$V} & \multicolumn{4}{c|}{abstract$\rightarrow$V} & \multicolumn{4}{c|}{intent$\rightarrow$V} & \multicolumn{4}{c}{any$\rightarrow$V}\\
\cline{2-17}
 Model  & R@1$\uparrow$ & R@5$\uparrow$ & R@10$\uparrow$ & MeanR$\downarrow$ & R@1$\uparrow$ & R@5$\uparrow$ & R@10$\uparrow$ & MeanR$\downarrow$ & R@1$\uparrow$ & R@5$\uparrow$ & R@10$\uparrow$ & MeanR$\downarrow$ & R@1$\uparrow$ & R@5$\uparrow$ & R@10$\uparrow$ & MeanR$\downarrow$ \\
  \hline
  InternVideo \cite{wang2022internvideo} &  82.10 &	95.50 &	97.50 	&2.40 &	77.60 &	92.50 &	95.50 &	4.10 &	50.80 &	72.10 &	78.30 &	49.70 &	70.13 &	86.69 	&90.42 &	18.74    \\
  UMT-L  \cite{li2023unmasked}& 87.56	&95.48&	96.86&	10.78	&78.35	&89.89	&93.16	&23.52&	37.39	&58.83	&65.61	&233.04  & 67.76&	81.40& 	85.21	&89.11\\
  CLIP-VIP \cite{xue2022clip} & 67.40 	&86.80 &	91.30 &	7.30 &	64.83&	83.90 	&88.43	&10.30 &	33.77&	55.27&	62.90 	&91.00 	&55.34	&75.32&	80.88	&36.02 \\
  mPLUG-2 \cite{xu2023mplug} & 83.20 &	93.20 &	95.43	&37.4	&77.57	&90.03&	93.23	&43.84	&35.17	&57.80 	&65.37	&235.72&	65.31&	80.34&	84.67&	105.65\\
          VAST \cite{chen2024vast}  &83.93	 &95.60  &	97.20  &	3.98	 &74.90  &	90.37 	 &92.87  &	9.69 &	43.83 &	66.07 &	72.87 &	73.05 &	67.55	 &84.01	 &87.64	 &28.90\\
% \textbf{DeME} & & & & \\
\hline
\end{tabular}
\end{table*}

\begin{table*}[t]
\caption{Results of hybrid-cognitive video-to-text retrieval on \textbf{VideoMind-3K}.}
\label{tab:v2t-deep-retrieval}
\vspace{-1em}
\small
\begin{tabular}{c|ccc|ccc|cccccc|cccc}
\hline
%   Model  & \multicolumn{4}{c|}{hybrid$\rightarrow$V} & \multicolumn{4}{c|}{factual$\rightarrow$V} & \multicolumn{4}{c|}{abstract$\rightarrow$V} & \multicolumn{4}{c}{intent$\rightarrow$V}\\
% \cline{2-17}
& \multicolumn{3}{c|}{Hit any layer} & \multicolumn{3}{c|}{Hit all layer}\\
\cline{2-7}
Model  & R@1$\uparrow$ & R@5$\uparrow$ & R@10$\uparrow$ & R@3$\uparrow$ & R@5$\uparrow$ &  R@10$\uparrow$ & TopR $\downarrow$ & LowestR $\downarrow$ &  AvgR $\downarrow$ \\
  \hline
  InternVideo \cite{wang2022internvideo} &79.63 &	93.77 	&96.23 &	30.67 &	45.10 &	58.80 &	2.75	&114.26&	43.79  \\
  UMT-L  \cite{li2023unmasked}&82.70	&93.00	&95.60	&29.73	&43.23&	54.23	&10.15&	523.49	&201.01\\
  CLIP-VIP \cite{xue2022clip} &  72.13&	88.97&	93.01	&19.23	&32.03&	44.40& 	7.21	&205.22&	79.99 \\
  mPLUG-2 \cite{xu2023mplug} & 72.97&	88.13 &	91.67	&26.26	&38.60 &	51.27	&23.51	&621.33&	254.96 \\
          VAST \cite{chen2024vast} &  78.83 &	92.10 &	95.17 	 &18.93 &	29.17	 &40.10  &	5.52	 &324.83 &	125.89\\
% \textbf{DeME}  & & & & \\
\hline
\end{tabular}
\end{table*}

%\section{Remarkable Performance on Downstream Tasks}

\section{Hybrid-cognitive Cross-modal Retrieval}
\textbf{VideoMind} provides multi-layer (from shallow to deep) textual descriptions of videos, and thus it is a comprehensive dataset for hybrid-cognitive video understanding. Particularly, it is the \textbf{FIRST} to support deep cognition of videos.
Cross-modal retrieval is selected as the primary downstream evaluation task. 
As described in Section \ref{sec:videomind}, 3,000 data samples were selected to form a benchmark for deep-cognitive cross-modal retrieval. The COT-generated texts of these data were meticulously validated by expert annotators. These data are strictly restricted to be used only for testing and \textbf{MUST NOT} be involved in any stage of model training.
In this section, we present the cross-modal retrieval results of several standard video-centric foundation models, including InternVideo \cite{wang2022internvideo}, UMT-L \cite{li2023unmasked}, CLIP-VIP \cite{xue2022clip}, mPLUG-2 \cite{xu2023mplug}, and VAST \cite{chen2024vast}. 
%Additionally, we report the results of the proposed \textbf{DeME} model, which is pre-trained using 10K training samples from \textbf{VideoMind}. This provides a novel benchmark for deep-cognitive cross-modal retrieval.

For text-to-video retrieval, we evaluate the retrieval capability of the models by utilizing texts at each layer, as well as the overall performance achieved through video search using texts from an arbitrary layer.
Rank K (R@K, $K\in\{1,5,10\}$) and Mean Rank (MeanR) are selected as evaluation metrics.
As anticipated, Table \ref{tab:t2v-deep-retrieval} demonstrates that the majority of foundation models attain satisfactory performance—reaching up to 87\% Rank-1 accuracy—when retrieving videos based on factually descriptive textual queries.
However, the retrieval capability diminishes progressively as the depth of expression increases. Specifically, the Rank-1 retrieval accuracy decreases to approximately 75\% when abstract-level texts are used as queries, and the performance deteriorates further ($\sim$35\%) when the intent layer is applied for retrieval.
Due to a lack of deep comprehension of video content, current foundation models demonstrate limited capability in representing the latent purposes of videos. Consequently, it becomes exceedingly difficult to retrieve accurate and relevant videos when using intent-layer textual queries that emphasize potential purposes rather than specific objects. It is observed that the MeanR can even exceed 200, indicating a significant misinterpretation of the video content.

For video-to-text retrieval,  the text expression at each layer is decomposed into three independent retrieval targets. we, respectively, compare two types of retrieval performance: 1) The success rate of retrieving any layer of the ground truth within the top K ($K\in\{1,5,10\}$) results; 2) The proportion of cases where all layers of the ground truth are included in the top K ($K\in\{3,5,10\}$) retrieval results.
We further record the earliest, latest, and average positions at which the three ground truth texts are retrieved, abbreviated as TopR, LowestR, and AvgR, respectively.
In Table \ref{tab:v2t-deep-retrieval}, the results in the "Hit any layer" column are relatively acceptable. The Rank-1 retrieval accuracy is around 80\% for most of these foundation models. This is because they can, at the very least, relatively easily search for the corresponding factual descriptions. Therefore, the highest search rank (TopR) is relatively acceptable ($\sim$ 10), which is always touched by the factual layer.
In contrast, all the tested models perform poorly in the "Hit all layer" metric. They are scarcely able to retrieve texts from all three layers within the top 10 results. Correspondingly, the LowestR can exceed 500, ranking nearly one-sixth among the 3,000 test data. 

Results in Tables \ref{tab:t2v-deep-retrieval} and \ref{tab:v2t-deep-retrieval} confirm the insufficiency of existing foundation models in in-depth video understanding.
As data-driven models, their limitations can be largely attributed to the unavailability of observable samples that convey deep semantic content. \textbf{VideoMind}, a dataset comprising 100,000 samples accompanied by deep-and-broad textual descriptions, is well-positioned to effectively address this gap.
% More experimental results are coming soon.

\section{Declaration}
 This is the first version of \textbf{VideoMind}. Our objective is to construct a million-scale video deep understanding database, which will serve as a foundational resource for advancing research and development in the field of video deep understanding. We believe \textbf{VideoMind} will accelerate the progress of video embeddings and various perception-related recognition tasks. More data will be continuously updated on various platforms. Please stay tuned for further developments.

%
% The next two lines define the bibliography style to be used, and
% the bibliography file.
\bibliographystyle{ACM-Reference-Format}
\bibliography{sample-base}

\clearpage

%%
%% If your work has an appendix, this is the place to put it.

\end{document}